\documentclass[journal,transmag]{IEEEtran}

\usepackage{graphicx}
\usepackage{algorithm}
\usepackage{algorithmic}
\usepackage[caption=false,font=footnotesize]{subfig}
\usepackage{longtable,array,ragged2e}
\newcolumntype{P}[1]{>{\RaggedRight\arraybackslash\hspace{0pt}}p{#1}}
\usepackage{amsthm}

\usepackage{url}
\usepackage{amsmath}
\usepackage[cmintegrals]{newtxmath}
\usepackage[table]{xcolor}
\usepackage{supertabular}
\usepackage{makecell}
\usepackage{multicol}

\usepackage[]{amssymb}
\usepackage[dvips]{epsfig}
\usepackage{lscape}
\usepackage{multirow}
\usepackage{color}
\usepackage{diagbox}
\usepackage{array}
\usepackage[normalem]{ulem}
\usepackage{ltxtable}
\usepackage{tabularx}
\usepackage{rotating}
\usepackage{verbatim}

\newcommand{\ds}[1]{\textcolor{blue}{#1}}

\newcommand\so{\bgroup\markoverwith{\textcolor{red}{\rule[0.5ex]{2pt}{0.4pt}}}\ULon}

\usepackage{xcolor}
\usepackage{colortbl}

\begin{document}

\title{Learn to Unlearn: Insights into Machine Unlearning}


\author{Youyang Qu,~\IEEEmembership{Member,~IEEE,}
        Xin Yuan,~\IEEEmembership{Member,~IEEE,}
        Ming Ding,~\IEEEmembership{Senior Member,~IEEE,} \\
        Wei Ni,~\IEEEmembership{Senior Member,~IEEE,} 
        Thierry Rakotoarivelo,~\IEEEmembership{Senior Member,~IEEE,} \\
        and David Smith,~\IEEEmembership{Senior Member,~IEEE}
\IEEEcompsocitemizethanks{\IEEEcompsocthanksitem Youyang Qu, Xin Yuan, Ming Ding, Wei Ni, Thierry Rakotoarivelo, and David Smith are with Data61, Commonwealth Scientific and Industrial Research Organization, Australia. Email: \{youyang.qu, xin.yuan, ming.ding, wei.ni, thierry.rakotoarivelo, smith.david\}@data61.csiro.au.

}\protect\\
}

\markboth{IEEE Computer Magazine}%
{Shell \MakeLowercase{\textit{et al.}}: Bare Demo of IEEEtran.cls for IEEE Transactions on Magnetics Journals}
%



\IEEEtitleabstractindextext{%
\begin{abstract}

Machine Learning (ML) models have been shown to potentially leak sensitive information, 
thus raising privacy concerns in ML-driven applications.
This inspired recent research on removing the influence of specific data samples from a
trained ML model.
Such efficient removal would enable ML to comply with the ``right to be forgotten'' in
many legislation, and could also address performance bottlenecks from low-quality or
poisonous samples. 
In that context, machine unlearning methods have been proposed to erase the contributions of
designated data samples on models, as an alternative to the often impracticable approach of
retraining models from scratch. 
This article presents a comprehensive review of recent machine unlearning techniques,
verification mechanisms, and potential attacks. 
We further highlight emerging challenges and prospective research directions (e.g. resilience
and fairness concerns). 
We aim for this paper to provide valuable resources for integrating privacy, equity, and
resilience into ML systems and help them ``learn to unlearn''.

\end{abstract}


}

\maketitle

\IEEEdisplaynontitleabstractindextext

%
\IEEEpeerreviewmaketitle

\section{Introduction}

\IEEEPARstart{I}{n} recent years, 
the proliferation of data and computational resources has led to the widespread adoption of Machine Learning (ML) models across various domains. 
From healthcare to finance, 
these models play an instrumental role in decision-making processes and predictive analytics. 
However, 
in the ever-changing landscape of data and user requirements,
static ML models, 
which remain unaltered after their initial training, 
become inadequate and ineffective. 
Consequently, 
a new paradigm emerges featuring a shift towards more flexible ML models.

On a separate front of the recent worldwide legislation on the rights of data privacy,
it becomes necessary to retrain ML models on datasets that have been modified. 
For instance, 
trained ML models should be modified when data points must be removed or amended~\cite{chien2023efficient}
due to privacy concerns, resilience requirements, 
bias reduction, 
or uncertainty quantification~\cite{wu2020deltagrad}.

In order to exclude certain data samples from a trained ML model, 
a new concept called machine \emph{unlearning} has been proposed recently to efficiently re-train an ML model without significantly sacrificing the ML performance. 
This concept opens up an alternative avenue to the traditional way of retraining the ML model entirely.

Machine unlearning is an emerging frontier in the realm of artificial intelligence and machine learning, 
focusing on the ability of models to forget or remove specific data or knowledge. 
At its core, 
this research topic addresses concerns related to data privacy, system updates, and model robustness. 
Traditional machine learning systems are designed to accumulate and retain information, 
but with the increasing scrutiny of data privacy and the dynamic nature of data sources, 
there's a pressing need for systems that can adapt by 'unlearning' outdated or undesired information. 
The proposed approaches in machine unlearning vary, 
ranging from retraining models from scratch to more sophisticated techniques that selectively prune or adjust the model's knowledge.
These approaches not only ensure compliance with data regulations but also pave the way for more responsible, flexible, and adaptive AI systems that can evolve with changing data landscapes.


Machine unlearning can 
primarily be categorised into ``exact'' and ``approximate'' unlearning paradigms.
Exact machine unlearning ensures the complete erasure of specific data from a model, 
rendering the model as if the data had never been introduced. Conversely, 
approximate unlearning aims for more efficient removal, 
albeit less precise, 
ensuring the model behaves similarly to one that has never seen the deleted data, 
though without guaranteeing exact equivalency. 

While both approaches offer promising pathways to address data privacy and model adaptability, 
the dichotomy between them introduces intricate challenges in verification and potential vulnerabilities to adversarial attacks. 
As these methodologies become foundational in modern AI systems, 
the imperative for research into their verification becomes paramount.
Ensuring the integrity and security of unlearning processes is not just a technical necessity but a mandate for the ethical deployment of AI in sensitive and dynamic data environments.

With the growing significance of machine unlearning, 
This paper strives to provide a concise yet comprehensive overview of the current landscape in this field. 
Our goal is to offer a systematic perspective on the ongoing advancements in this area and to highlight its critical role in the constantly evolving machine learning ecosystem.

The main contributions of this article are summarized as follows.
\begin{itemize}
    \item A comprehensive survey of existing research on exact and approximate machine unlearning, 
    with a taxonomy presented in Table~\ref{table:accuracy}.
    \item A thorough examination of the verification of machine unlearning and relevant attacks.
    \item Experimental results that provide valuable insights into the effectiveness and cost of machine unlearning.
\end{itemize}

In this study, we focused on representative research that encompasses both exact and approximate machine unlearning techniques, in addition to delving into associated attacks and verification methods. To ensure a thorough analysis of recent advancements in the field, we selected research papers spanning from 2015 through 2022, from top-ranked peer-reviewed publication venues in the IEEE and ACM databases.

\section{From learning to unlearning}

\subsection{Motivation}


Privacy concerns, security, and performance optimization underpin the need for machine unlearning in the broader landscape of machine learning models and their applications.

Firstly, from a privacy protection perspective, 
the European Union (EU) implemented the General Data Protection Regulation (GDPR) in 2016, which is
widely recognized as the most stringent legislation regarding data privacy and security globally~\cite{voigt2017eu}. 
This regulation enforces obligations on organizations worldwide that collect or use data relating to individuals within the EU. 
It recognizes novel data privacy rights to grant individuals greater autonomy over their data, 
including the right to erasure (i.e., {\em right to be forgotten}).

In 2018, 
California passed the California Consumer Privacy Act (CCPA)\footnote{https://oag.ca.gov/privacy/ccpa},
providing consumers with substantial control over the personal information that businesses collect about them. The ``right to delete data'' defined in the CCPA is the equivalent of the right to erasure in the GDPR.
Recently, 
California established the California Privacy Rights Act (CPRA)\footnote{https://oag.ca.gov/privacy/ccpa}, 
also known as ``CCPA 2.0''. 
This Act extensively expands and modifies the CCPA, rendering it comparable to the GDPR. 
The CPRA came into effect in January 2023. 
Over the same period, the Australian Privacy Act Review Report\footnote{https://www.ag.gov.au/rights-and-protections/publications/privacy-act-review-report} has recommended adding the ``right to be forgotten'' to Australian legislation. 

Secondly, in the realm of cybersecurity, machine unlearning is crucial for poisoning attack recovery. This involves the removal of maliciously injected, poisoned data samples, which could otherwise adversely affect the model's predictive accuracy or behaviour. 

Lastly, from a performance optimization perspective, machine unlearning aids in eliminating the influence of low-quality data samples. These may have been incorporated during the training process and could have degraded the overall performance of the model. Hence, unlearning these specific instances can enhance the model's efficiency and effectiveness~\cite{cao2015towards,guo2020certified,nguyen2020variational}. Therefore, machine unlearning emerges as an essential aspect of managing and improving ML models in contemporary data-driven applications.

\begin{figure}
\centering
\includegraphics[width=3.4in]{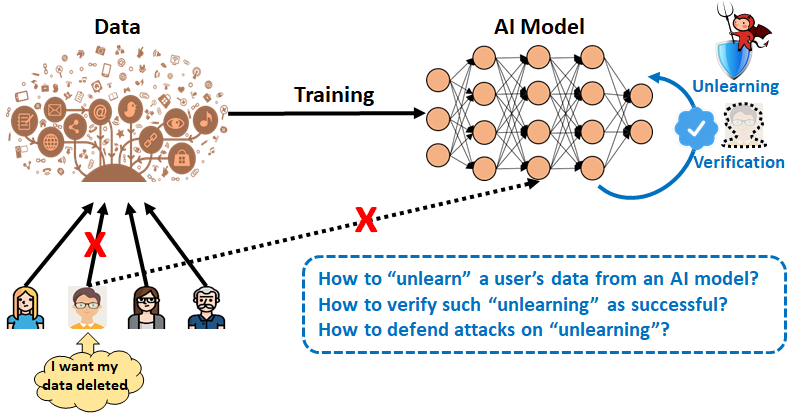}
\caption{Diagram of a \ds{general scenario for} machine unlearning}
\label{diagram}
\end{figure}

\subsection{Overview of Machine Unlearning}

Fig.~\ref{diagram} illustrates the general scenario for Machine unlearning. At its core, it is the art of selective forgetting. It revolves around the principle of inverse operations, which aims to mathematically reverse or negate the effect of certain data on a model's parameters. Every data point that a machine learning model processes induces specific modifications to its parameters, and recognizing these nuances becomes pivotal in the unlearning process.

Several algorithms have been proposed to facilitate this intricate process. The straightforward method involves the complete removal of undesired data followed by retraining, ensuring the model is purged of the specific knowledge. Another approach is weight adjustment, where the influence of the unwanted data is counteracted by directly tweaking the model's weights, negating the need for complete retraining. Leveraging regularization methods, some algorithms impose penalties on the model for retaining specific data, coaxing it to forget. Differential unlearning, a more recent technique, focuses on understanding the changes in model weights induced by particular data points during training and then reversing these changes to achieve unlearning.

Supporting these algorithms are several mechanisms that enhance the efficiency and reliability of the unlearning process. Model versioning, for instance, maintains snapshots of various model states, allowing for quick reversion to a prior state when unlearning becomes necessary. Another significant development is the integration of data sanitization layers within the training process. These layers continuously screen and filter out data that's flagged for unlearning. Additionally, checkpoint systems, which store periodic model states, offer rapid rollback options, enabling efficient unlearning without trudging through the entire training data again.

In essence, as the digital landscape becomes increasingly dynamic, machine unlearning, underpinned by these foundational concepts, algorithms, and mechanisms, ensures our models remain adaptable and up-to-date.

\section{Exact Machine Unlearning}

Exact ML methods aim at eliminating  
the influence of the particular segment/proportion of data to be removed completely. Generally, some degree of retraining is necessary. The major challenges encountered with this approach
involve retraining efficiency and the subsequent performance of the retrained model.

The pioneering study in the domain of machine unlearning was conducted by Cao \textbf{et al.}~\cite{cao2015towards}, marking the initial introduction of this concept. Their research entailed the proposal of a heuristic methodology that allowed for the transformation of an ML algorithm into a summation format, effectively neutralizing the influence of data lineage. The non-uniform distribution of unlearning requests in practical applications was also considered within their study. However, it merits acknowledgment that not every ML algorithm can undergo conversion into the summation format, and the data lineage varying in granularity levels might not always be addressed.

To enhance the precision of machine unlearning techniques, Bourtoule \textit{et al.}~\cite{bourtoule2021machine} introduced an alternative method referred to as SISA. This approach aims to ameliorate the universality and performance of the retrained model. The acronym SISA signifies ``Sharded, Isolated, Sliced, and Aggregated training'', and it realizes machine unlearning via incremental unlearning on partitioned data. This technique involves the categorization of data into several isolated shards, followed by further segmenting of the data within each shard. Subsequent to this, incremental learning is implemented, and parameters are archived. Upon receiving a new unlearning request, SISA navigates back to the relevant data slice and initiates retraining from that point. Given that the remaining model parameters have already been preserved, the process of aggregation becomes straightforward and efficient. Nonetheless, SISA could potentially experience a drop in performance in each component model when datasets are not sufficiently large, or the learning task is particularly intricate.

\begin{figure*}[!htbp]
\centering
\subfloat[Gradient Descend (Loss Minimization)]{
\includegraphics[width=3.2in]{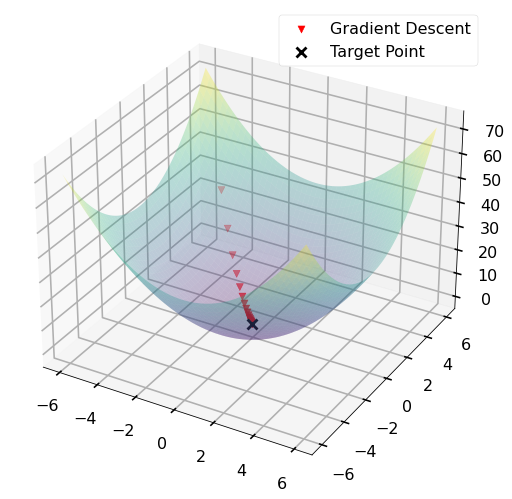}
\label{f_2a}
}
\hfil
\subfloat[Gradient Ascent (Loss Maximization)]{
\includegraphics[width=3.2in]{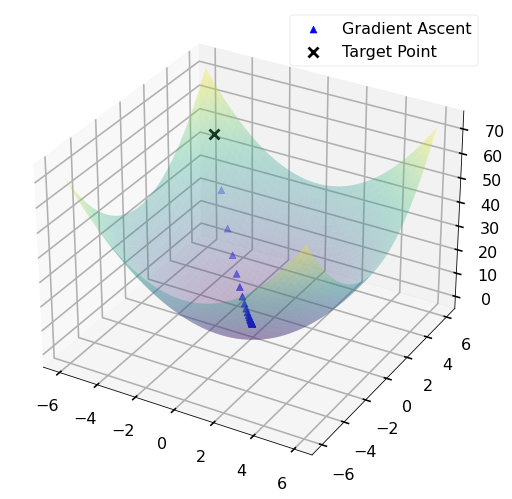}
\label{f_2b}
}
\caption{Machine Learning (Loss Minimization) v.s. Machine Unlearning (Loss Maximization)}
\label{illusration}
\end{figure*}

``DeltaGrad''~\cite{wu2020deltagrad}, an exact machine unlearning methodology, emphasizes expedited retraining, as shown in Fig.~\ref{illusration}. It accomplishes this by training the data designated for deletion in a counteractive manner – that is, by maximizing the loss rather than minimizing it, as is typical in traditional ML scenarios. The resulting reversed model is subsequently integrated with the original model, which is preserved beforehand.
Through the application of DeltaGrad, the performance of the model can be well-sustained, given certain constraints on the proportion of data to be removed (1\% as suggested in the DeltaGrad publication). Nevertheless, although DeltaGrad is compatible with stochastic gradient descent ML algorithms, it lacks the capability to manage mini-batch sizes.

The gradient descent method of machine learning is depicted in  Fig.~\ref{f_2a}. The red arrows descend towards a target point, minimizing the loss function. This is the traditional optimization process of training an ML model. In contrast, machine unlearning, as shown in Fig.~\ref{f_2b}, follows a different trajectory. The blue arrows ascend towards a target point, maximizing the loss function. This process seeks to emphasize and amplify the errors associated with the designated data for deletion, ultimately removing their influence from the ML model.

Rather than general-purpose machine unlearning methodologies, focused strategies tailored to specific ML algorithms are currently being devised. To illustrate, Brophy \textbf{et al.}~\cite{brophy2021machine} presented ``DaRE'', a technique specifically engineered for the random forest algorithm. Even though a minor performance reduction in the model is observed, DaRE is capable of swiftly retraining, enhancing efficiency by 2 to 4 orders of magnitude compared to completely starting anew. However, the application of DaRE is limited solely to the random forest algorithm and does not extend to other ML algorithms.

In conclusion, there has been considerable investigation into exact machine unlearning from diverse perspectives. Till now, a majority of the research has centered on rapid retraining techniques, with minimal or ideally no compromise in the performance of the model. Theoretically, the removed data can be validated through the retrained model, given that the primary objective of exact machine unlearning is to physically eliminate the influence of the excised data without disrupting the retrained model.

\textbf{Discussion}: Although exact machine unlearning techniques have made substantial advancements towards actualizing the right to erasure, researchers have pinpointed emerging challenges from two distinct viewpoints. Firstly, exact machine unlearning invariably induces a deterioration in model performance, albeit occasionally the decay may be marginal. Secondly, the prospect of privacy leakage is a concern. For instance, should a data entity named ``Alice'' be eliminated and a discrepancy is detected between the original and the retrained model, a hostile entity could infer that the observed difference is attributable to the removal of ``Alice''. Consequently, this could instigate an inference attack, aggravating the privacy risk.

\section{Approximate Machine Unlearning}

To mitigate the challenges of exact machine unlearning, approximate machine unlearning methods have been developed, which try to mask the difference between models before and after data removal while simultaneously optimizing model performance.

A prominent method in the realm of approximate machine unlearning is the ``certified-removal'' approach as proposed by Guo \textit{et al.}~\cite{guo2020certified}. Their research illustrates a mechanism to expunge data points from $L_2$-regularized linear models, provided these models have been trained using a differentiable convex loss function. The approach deployed by them utilizes a Newton step on the model parameters, which proficiently eradicates the influence of the data point under deletion. Additionally, the residual is obscured by introducing random perturbations into the training loss, thereby preventing potential privacy violations. Despite its groundbreaking nature, the proposed methodology has certain restrictions. Firstly, implementing Newton's optimization method requires the inversion of the Hessian matrix, a task that can pose considerable challenges. Secondly, the technique is not adaptable to models with non-convex losses. Lastly, the data-dependent bound does not quite succeed in accurately estimating the gradient residual norm, which underlines the need for further enhancements and rigorous scrutiny.

To explore approximate machine unlearning for intricate models, notably deep neural network (DNN) models, Golatkar \textit{et al.}~\cite{golatkar2020eternal} introduced a forgetting Lagrangian to accomplish selective forgetting in DNNs. Building upon this concept, the authors devised a scrubbing method capable of erasing information from trained weights without the need to access the original training data or mandate a complete retraining of the network. Moreover, they developed a computable upper bound to measure the amount of residual information, a value that can be efficiently computed for DNNs. Nevertheless, when it pertains to forgetting without assuming prior training, research has revealed that even slight perturbations during the crucial learning phase can engender substantial variations in the eventual solution.

The aforementioned two research contributions implemented differential privacy mechanisms to disturb the retrained model parameters, raising the inquiry of whether machine unlearning can be considered a special instance of differential privacy within the field of machine learning. Sekhari \textit{et al.}~\cite{sekhari2021remember} tackled this matter by undertaking an exhaustive exploration of generalization in machine unlearning, with the objective of achieving commendable performance on new data points. In contrast to earlier research, the algorithms deployed in this study do not require the unlearning algorithm to have access to the training data during the deletion of samples. This study delineates a clear distinction between differential privacy and machine unlearning; however, it does contain a few limitations. Specifically, it does not provide dimension-dependent information-theoretic lower bounds on the deletion capacity and is unable to handle non-convex loss functions.

In contrast to the perturbation-based methodologies, Nguyen \textit{et al.}~\cite{nguyen2020variational} proposed a machine unlearning technique for variational Bayesian machine learning, a significant sub-discipline of machine learning. They accomplished this by integrating Bayesian theory into the model's prior and posterior knowledge, thereby extending machine unlearning into the Bayesian domain. Though employing stochastic methodologies, the authors approached the issue from a probabilistic theory perspective, as opposed to the randomized noise injection method employed in differential privacy. This unlearning technique does not compromise the model's performance and is exclusively applicable to this specific segment of ML models.

In the field of approximate machine unlearning, there are solutions purpose-built for specific ML models. For example, Ginart \textit{et al.}~\cite{ginart2019making} have developed machine unlearning techniques that are custom-designed for K-means. The primary objective of these methods is to strike an equilibrium between efficiency and model performance. It should be noted that these methods are constrained to eliminating merely one record per iteration. Moreover, the authors formulated four fundamental principles for the design of efficient machine unlearning methodologies. These guiding principles include linearity, laziness, modularity, and quantization.


\textbf{Discussion}: In summary, approximate machine unlearning places a higher premium on preserving model performance while introducing additional privacy safeguards to the retrained model. However, it confronts several challenges, including the intricacy of verifying the implementation of machine unlearning methodologies. In fact, based solely on output observations, it becomes nearly impossible to distinguish whether the unlearning process has been enacted or not. Furthermore, since the majority of approximate machine unlearning models initiate with the model rather than the data, it prompts questions regarding the congruity of this method with the ``right to be forgotten'', as stipulated in privacy laws and regulations.

\clearpage
\onecolumn

\begin{table*}
    \caption{Taxonomy of Machine Unlearning}
    \label{table:accuracy}
\end{table*}
\begin{longtable}[l]{|l|>{\raggedright}p{1.3cm}|>{\raggedright}p{1.3cm}|>{\raggedright}p{2.1cm}|>{\raggedright}p{6.7cm}|>{\raggedright}p{3.9cm}|}
\hline 
 & \textbf{Reference} & \textbf{Dataset} & \textbf{Methodology} & \textbf{Performance Summary} & \textbf{Security \& Privacy} \tabularnewline
\hline 
\hline
\endhead
\hline 
\multirow{4}{*}[-5.5cm]{\begin{turn}{90}
\textbf{\Large{}Exact Unlearning}
\end{turn}} & {\footnotesize{}Cao et al. {[}4{]}} & {\footnotesize{}Collected from Facebook} & {\footnotesize{}Converting algorithms into a summation form for efficiently
forgetting data lineage} & \begin{itemize}
\item {\footnotesize{} Efficiency: The idea of forgetting systems that can restore privacy,
security, and usability by completely and quickly erasing data lineage}{\footnotesize\par}
\item {\footnotesize{}Accuracy: The accuracy performance depends on the percentage of the deletion data}{\footnotesize\par}
\item {\footnotesize{}Limitations: Not all ML algorithm supports summation form, granularity issues}{\footnotesize\par}
\end{itemize}
 & \begin{itemize}
\item {\footnotesize{}Real-world data pollution attacks developed against
systems and algorithms}{\footnotesize\par}
\end{itemize}
\tabularnewline
\cline{2-6} \cline{3-6} \cline{4-6} \cline{5-6} \cline{6-6} 
 & {\footnotesize{}Bourtoule et al. {[}7{]}} & {\footnotesize{}Purchase, SVHN} & {\footnotesize{}SISA stands for Sharded, Isolated,
Sliced, and Aggregated training} & \begin{itemize}
\item {\footnotesize{}Efficiency: The incremental unlearning mode of SISA is efficient}{\footnotesize\par}
\item {\footnotesize{}Accuracy: A combination of transfer
learning and SISA yields a slight decrease in accuracy (\textasciitilde 2\%)
with improved retraining time.}{\footnotesize\par}
\item {\footnotesize{}Limitations: Hard to Verify and cannot deal with small datasets}{\footnotesize\par}
\end{itemize}
 & \begin{itemize}
\item {\footnotesize{}Can well-resist membership inference attacks}{\footnotesize\par}
\item {\footnotesize{}Vulnerable to poisoning attacks}{\footnotesize\par}
\end{itemize}
\tabularnewline
\cline{2-6} \cline{3-6} \cline{4-6} \cline{5-6} \cline{6-6} 
 & {\footnotesize{}Wu et al. {[}2{]}} & {\footnotesize{}MNIST, RCV1, HIGGS} & {\footnotesize{}Storing the original gradients, retaining the deleted
data by gradient ascend, and then integrating} & \begin{itemize}
\item {\footnotesize{}Efficiency: Can quickly (2-6 times quicker) retrain machine learning models
based on stochastic gradient ascend}{\footnotesize\par}
\item {\footnotesize{}Accuracy: Negligible changes in performance in response to small changes
in the data} {\footnotesize\par}
\item {\footnotesize{}Limitations: Extra storage, mini-batch not supported, requiring convexity of ML models} {\footnotesize\par}
\end{itemize}
 & \begin{itemize}
\item {\footnotesize{}The storage of all inter-media model parameters pose a threat to model inversion attacks}{\footnotesize\par}
\item {\footnotesize{}Gradient ascent for deletion samples may lead to risks to other similar records not to be deleted}{\footnotesize\par}
\end{itemize}
\tabularnewline
\cline{2-6} \cline{3-6} \cline{4-6} \cline{5-6} \cline{6-6} 
 & {\footnotesize{}Brophy et al. {[}8{]}} & {\footnotesize{}Adult, HIGGS, Bank, etc.} & {\footnotesize{}Machine unlearning for random forests} & \begin{itemize}
\item {\footnotesize{}Efficiency: Swift removal of training instances for a novel variant of random forests called DaRE, with a speed advantage of 2\textasciitilde 4 orders}{\footnotesize\par}
\item {\footnotesize{}Accuracy: Slight performance degradation (less than 1\%) based on in handling
sequences of deletions on 13 binary classification datasets from the
real world}{\footnotesize\par}
\item {\footnotesize{}Limitations: Only applicable to Random Forest} {\footnotesize\par}
\end{itemize}
 & \begin{itemize}
\item {\footnotesize{} Privacy leakage potential before and after deletion}{\footnotesize\par}
\end{itemize}
\tabularnewline
\hline 
\multirow{2}{*}[-1cm]{\begin{turn}{90}
\textbf{\Large{}Approx. Unlearning}
\end{turn}} & {\footnotesize{}Guo et al. {[}5{]}} & {\footnotesize{}MNIST} & {\footnotesize{}Training a differentiable convex loss
function in a certified manner that is from L2-regularized
linear models} & \begin{itemize}
\item {\footnotesize{} Efficiency: Utilize a Newton step to effectively remove the influence
of the removed data point on the model parameters}{\footnotesize\par}
\item {\footnotesize{} Accuracy: Indistinguishable performance when the deleting sample size is within a certain range}{\footnotesize\par}
\item {\footnotesize{} Limitations: Difficult to invert the Hessian matrix, non-convex losses not supported, large disparity} {\footnotesize\par}
\end{itemize}
 & \begin{itemize}
\item {\footnotesize{}Proposes to mask the residual by randomly perturbing
the training loss to prevent adversaries from extracting information
from small residual}{\footnotesize\par}
\end{itemize}
\tabularnewline
\cline{2-6} \cline{3-6} \cline{4-6} \cline{5-6} \cline{6-6}  
 & {\footnotesize{}Golatkar et al. {[}9{]}} & {\footnotesize{}MNIST, CIFAR-10, Lacuna-10} & {\footnotesize{}Defining a forgetting Lagrangian to achieve selective
forgetting for deep neural networks} & \begin{itemize}
\item {\footnotesize{}Efficiency: The selective forgetting method can improve the efficiency by careful design }{\footnotesize\par}
\item {\footnotesize{}Accuacy: If the pretraining assumption is not considered while
forgetting, even minor perturbations during the initial critical learning
phase can result in significant variations in the final solution}{\footnotesize\par}
\item {\footnotesize{}Limitation: Requires pre-training to guarantee convergence} {\footnotesize\par}
\end{itemize}
 & \begin{itemize}
\item {\footnotesize{}Propose a scrubbing procedure that removes information
from the trained weights without accessing the raw training data or
re-training from scratch} {\footnotesize\par}
\end{itemize}
\tabularnewline
\hline
\multirow{3}{*}[-4cm]{\begin{turn}{90}
\textbf{\Large{}Approximate Unlearning}
\end{turn}} & {\footnotesize{}Sekhari et al. {[}10{]}} & {\footnotesize{} Experimental evaluation not provided} & {\footnotesize{}Generalization in machine unlearning for performing
well on unseen data} & \begin{itemize}
\item {\footnotesize{}Efficiency: Explore the number of samples that can be unlearned
while preserving performance}{\footnotesize\par}
\item {\footnotesize{}Accuracy: Comparable performance of generalization in machine unlearning for performing
well on unseen data}{\footnotesize\par}
\item {\footnotesize{}Limitations: Fail to provide dimension-dependent information-theoretic
lower bound on the deletion capacity, cannot deal with non-convex loss functions}{\footnotesize\par}
\end{itemize}
 & \begin{itemize}
\item {\footnotesize{}Do not rely on the availability of training data during
sample deletion, distinguishing it from prior work}{\footnotesize\par}
\item {\footnotesize{} Show a strict separation between DP and machine unlearning}{\footnotesize\par}
\end{itemize}
\tabularnewline
\cline{2-6} \cline{3-6} \cline{4-6} \cline{5-6} \cline{6-6} 
 & {\footnotesize{}Ginart et al. {[}11{]}} & {\footnotesize{}MNIST, CellType, Botnet} & {\footnotesize{}Machine unlearning for K-means while synthesizing 4 general engineering principles} & \begin{itemize}
\item {\footnotesize{}Efficiency: Introduce and characterize the problem of efficient
data deletion in machine learning}{\footnotesize\par}
\item {\footnotesize{}Accuracy: Propose two solutions to achieve deletion efficiency
in k-means clustering, offering theoretical guarantees and strong
empirical performance}{\footnotesize\par}
\item {\footnotesize{}Limitations: Can only unlearn one single data point at one time, an assumption that one model corresponds to one dataset} {\footnotesize\par}
\end{itemize}
 & \begin{itemize}
\item {\footnotesize{}Maybe vulnerable to membership inference attacks by manipulating the center of K-means}{\footnotesize\par}
\item {\footnotesize{}The models are vulnerable to data sample addition/deletion} {\footnotesize\par}
\end{itemize}
\tabularnewline
\cline{2-6} \cline{3-6} \cline{4-6} \cline{5-6} \cline{6-6} 
 & {\footnotesize{}Nguyen et al. {[}6{]}} & {\footnotesize{}Airline, Moon, Fashion MNIST} & {\footnotesize{}Machine unlearning for variational Bayesian ML} & \begin{itemize}
\item {\footnotesize{}Efficiency: Extend machine unlearning to efficient and feasible Variational Bayesian
Unlearning}{\footnotesize\par}
\item {\footnotesize{}Accuracy: Comparable performance after evaluating unlearning methods on Bayesian models, sparse Gaussian process, logistic regression}{\footnotesize\par}
\item {\footnotesize{}Limitations: vulnerable to attacks, only applicable to variational inference frameworks of Bayesian machine learning}{\footnotesize\par}
\end{itemize}
 & \begin{itemize}
\item {\footnotesize{}New vulnerabilities of model fraud attacks}{\footnotesize\par}
\end{itemize}
\tabularnewline
\hline 
\multirow{3}{*}[-2.5cm]{\begin{turn}{90}
\textbf{\Large{}Verification \& Attacks}
\end{turn}} & {\footnotesize{}Sommer et al. {[}12{]}} & {\footnotesize{}MNIST, CIFAR-10, Fashion MNIST, etc.} & {\footnotesize{}Probabilistic verification of machine unlearning leveraging the backdoor} & \begin{itemize}
\item {\footnotesize{}Verification Efficiency: Acceptable verification efficiency by identifying the pre-injected backdoor}{\footnotesize\par}
\item {\footnotesize{}Verification Accuracy: Superior accuracy based on the verification performance using multiple datasets}{\footnotesize\par}
\item {\footnotesize{}Verification Limitations: Constraints on data samples, conflicting backdoor patterns, potential vulnerability to backdoor detection mechanisms}{\footnotesize\par}
\end{itemize}
 & \begin{itemize}
\item {\footnotesize{}High risk of backdoor injection for attacks instead of verification}{\footnotesize\par}
\end{itemize}
\tabularnewline
\cline{2-6} \cline{3-6} \cline{4-6} \cline{5-6} \cline{6-6} 
 & {\footnotesize{}Chen et al. {[}13{]}} & {\footnotesize{}MNIST, CIFAR-10, Adult, etc.} & {\footnotesize{}Proposed member inference attacks on machine unlearning} & \begin{itemize}
\item {\footnotesize{}Attack Efficiency: Can efficiently launch the proposed member inference attack}{\footnotesize\par}
\item {\footnotesize{}Attack Success Ratio: The attack can easily breach the privacy of ML model-related unlearning but not the data-related unlearning like SISA}{\footnotesize\par}
\item {\footnotesize{}Limitations: Not applicable to data-manipulation-based unlearning methods, performance drops if the unlearned/updated samples represent more than 0.2\%} {\footnotesize\par}
\end{itemize}
 & \begin{itemize}
\item {\footnotesize{}Priavcy-breacching attack to identify the existence of a record}{\footnotesize\par}
\item {\footnotesize{}Quantify the privacy risks in machine unlearning by
examining membership inference attacks}{\footnotesize\par}
\end{itemize}
\tabularnewline
\cline{2-6} \cline{3-6} \cline{4-6} \cline{5-6} \cline{6-6} 
 & {\footnotesize{}Marchant et al. {[}14{]}} & {\footnotesize{}MNIST, Fashion- MNIST, etc.} & {\footnotesize{}Poisoning attacks on certified machine unlearning} & \begin{itemize}
\item {\footnotesize{}Attack Efficiency: Requires multiple rounds of poisoning operation with a balance of the attacker's objective optimality with the computation time and the feasibility of long-term attack execution}{\footnotesize\par}
\item {\footnotesize{}Attack Success Ratio: Consider different factors such as white-box and grey-box attacks, various perturbation geometries and bounds, etc}{\footnotesize\par}
\item {\footnotesize{}Limitations: Only targets on a particular category of machine unlearning
models, not work well if off-the-shelf anomaly detection models have been deployed}{\footnotesize\par}
\end{itemize}
 & \begin{itemize}
\item {\footnotesize{}Significant threats to certified machine unlearing}{\footnotesize\par}
\item {\footnotesize{}Poisoning strategy can evolve with unlearning process}{\footnotesize\par}
\end{itemize}
\tabularnewline
\hline 
\end{longtable}

\pagebreak{}

\clearpage
\twocolumn

\section{Verification and Attacks of Unlearning Algorithms}
\label{sec::verification}

Concurrent with the design of novel unlearning methods, mechanisms for systematically verifying data erasure and attacks on machine unlearning have been developed. During our survey, we discovered only one research contribution related to the former~\cite{sommer2022athena}, with a handful of other contributions related to the latter~\cite{chen2021machine,marchant2022hard}.

Regarding the verification mechanism, Sommer \textit{et al.}~\cite{sommer2022athena} have conducted research on probabilistic verification of machine unlearning. In cases where data needs to be erased, data owners intentionally implant a backdoor within the data prior to transmitting it to the data user (for instance, Internet titan corporations). Once the data user claims to have deleted the data, the data owner can corroborate the erasure by checking the previously inserted backdoor. In addition, there exist several strategies aiding in the verification of machine unlearning, encompassing but not limited to data sampling verification, synthetic data verification, and bias analysis. For instance, bias analysis involves assessing the effect of data elimination on the bias inherent in the ML model. This could entail gauging the fairness and accuracy of the model before and after data removal or examining the data distribution to discern the influence of data removal on the representation of various groups. The mechanism has demonstrated its effectiveness across an array of datasets and networks. Nevertheless, a few limitations include restrictions on data samples, conflicting backdoor patterns, risks associated with backdoor injection, and potential susceptibility to backdoor attack detection mechanisms.

Membership inference is a type of privacy-oriented attack proposed against ML models~\cite{shokri2017}, and it bears relevance to machine unlearning. In their recent work, Chen \textit{et al.}~\cite{chen2021machine} unveiled a novel method for executing membership inference attacks within the context of machine unlearning. This approach allows them to ascertain whether a particular sample was part of the training set of the original model, thereby illuminating unforeseen privacy risks tied to machine unlearning. However, the success rates of the attack decrease under specific conditions, such as the application of the SISA method and when more than 0.2\% of data samples require deletion.

Beyond the attacks discussed earlier, poisoning attacks are a prevalent form of assault in ML applications, inclusive of machine unlearning. Marchant \textit{et al.}~\cite{marchant2022hard} have conceived an innovative poisoning attack that accounts for various factors. These include white-box and grey-box attacks, diverse perturbation geometries and limits, optimization of the attacker's objective harmonized with computational time, and the feasibility of sustaining the attack over an extended duration. Nonetheless, this attack is tailored specifically for a unique class of machine unlearning models, specifically, the certified-removal~\cite{guo2020certified} and its variants. Its performance may deteriorate if standard anomaly detection models have been implemented.

\section{Evaluation and Experimental Results}
\label{sec::summary}

In the developing field of machine unlearning, there are discernible gaps and ambiguities in the existing literature. These gaps not only leave pressing questions unanswered but also hint at the potential for fresh perspectives that might reshape our current understanding. Our series of experiments, presented in this survey, are not merely a reiteration of known concepts but an endeavour to introduce novel ideas and concepts that have been largely unexplored in contemporary discourse. By weaving these empirical findings into our review, we aim to break the traditional paradigm of survey papers, offering readers not just a synthesis of existing knowledge but also tangible evidence of recent advances. This integration of experiments serves a dual purpose: it provides a more comprehensive view of machine unlearning and ensures an engaging read by interspersing theoretical discussions with practical insights.

Specifically, 
our experiments are based on Deltagrad~\cite{wu2020deltagrad}, 
which is considered to be one of the most prominent and precise machine unlearning techniques. 
As previously mentioned, 
the key approach employed in this study is ``reverse learning''. 
To elaborate further, 
Deltagrad amplifies the loss of the excluded data samples in the model training, 
instead of using the conventional method that minimizes the loss.

\subsection{Experiment Design}
Our evaluation procedure of machine unlearning is as follows. 
Firstly, we explore the relationship between the accuracy of the model and the percentage of deleted data samples in the independent and identically distributed (IID) scenario. 
Secondly,  we investigate the same correlation in the non-IID situation. 
Thirdly, we assess the time consumption associated with the proportion of the deleted data samples.

For our experiments, we used the original ``Deltagrad''\footnote{https://github.com/thuwuyinjun/DeltaGrad} implementation and further extended it with new features to accommodate the evaluation of non-IID deletion and accuracy of each class of different datasets\footnote{https://github.com/DrQu89757/Extened\_code\_from\_DeltaGrad}. These evaluations and related results were not considered in the original ``Deltagrad'' article.
All presented results are the mean values over 20 rounds of experiments, with error bars showing the corresponding standard deviations.

\begin{figure*}[!htbp]
    \centering
    \subfloat[IID]{
    \includegraphics[width=3.2in]{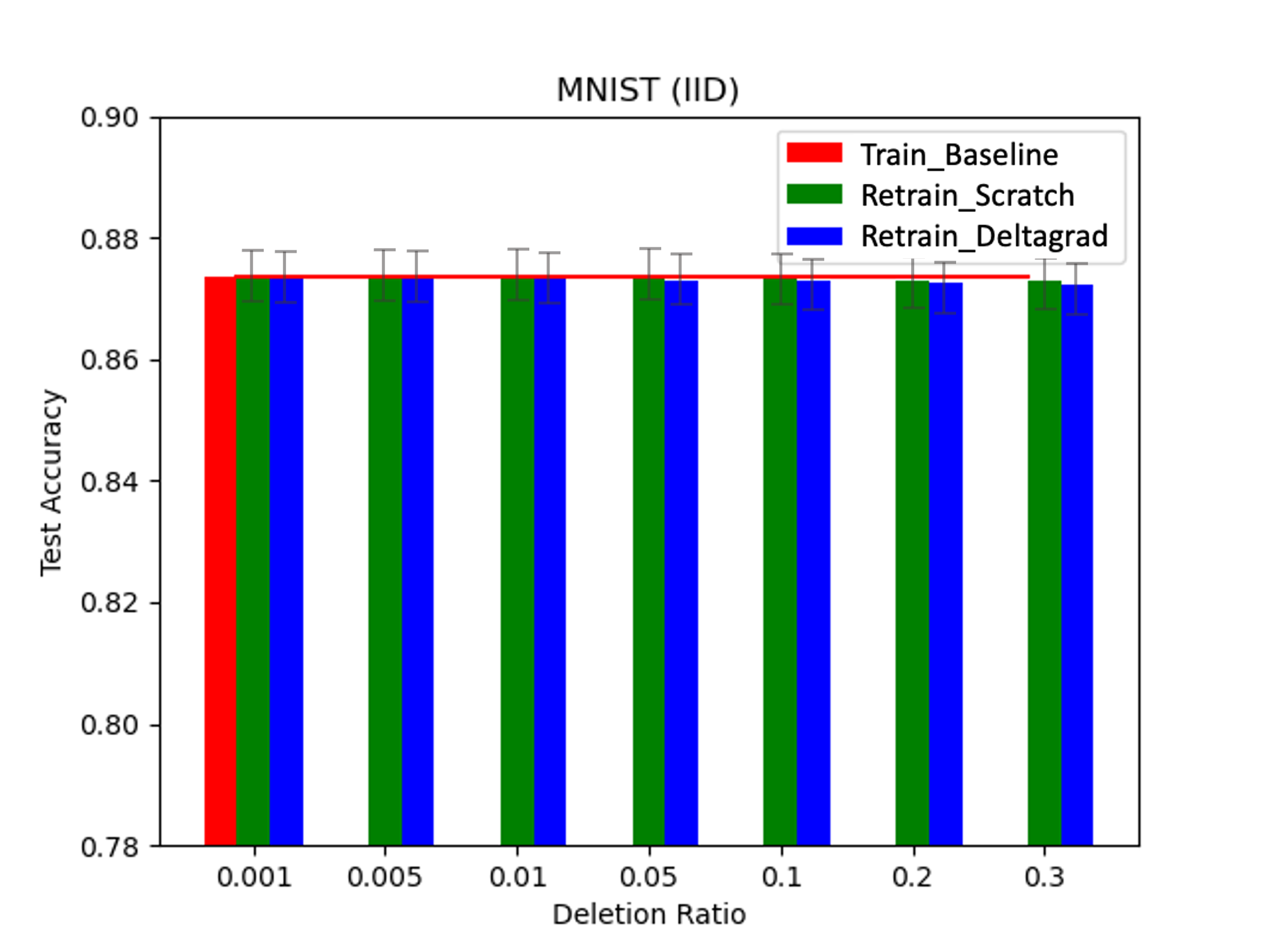}
    \label{fig_iid}
    }
    \hfil
    \subfloat[non-IID]{
    \includegraphics[width=3.2in]{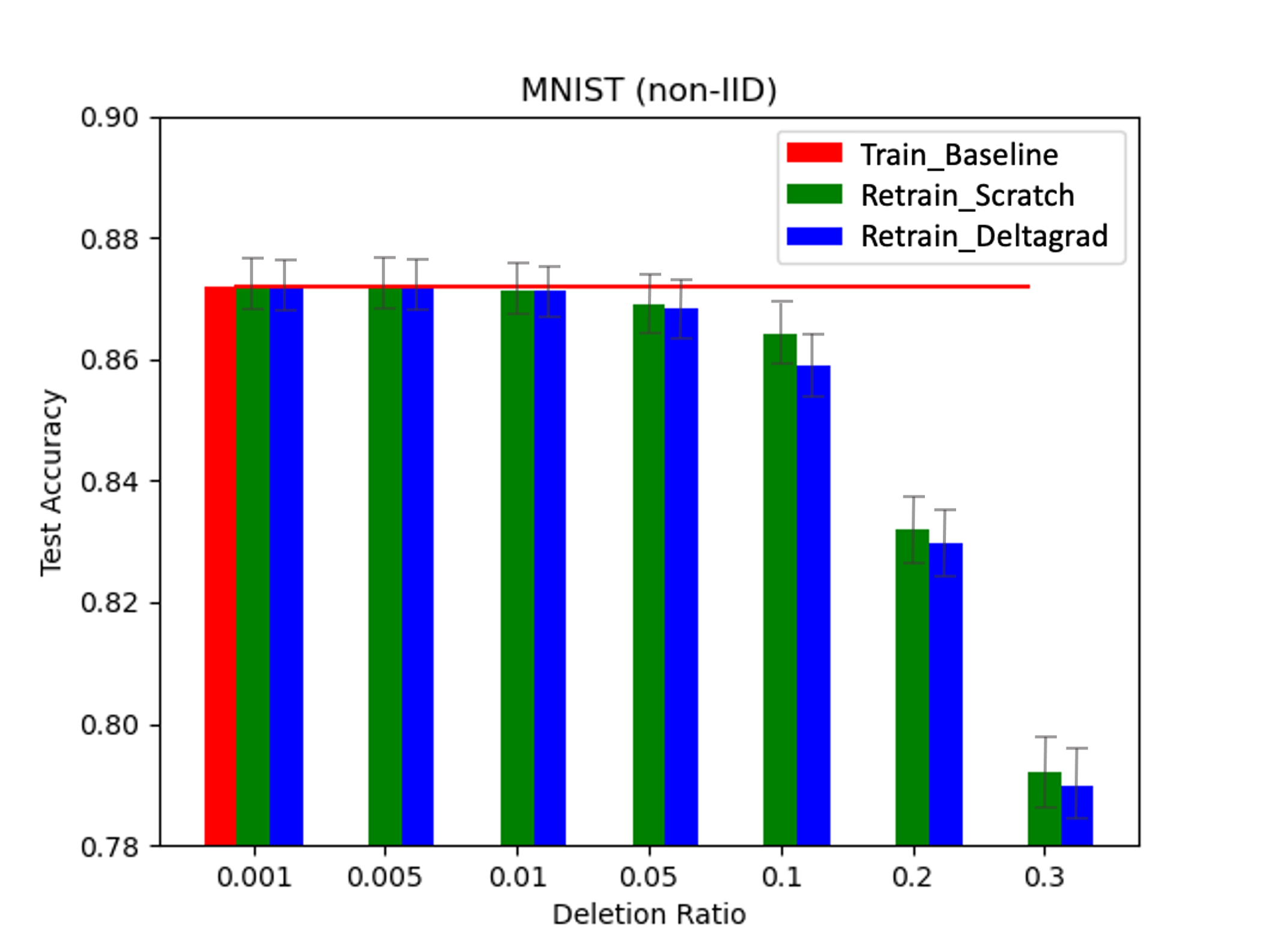}
    \label{fig_noniid}
    }
    \caption{Evaluation of accuracy in IID and non-IID settings}
    \label{fig:fig_7}
\end{figure*}

\subsection{Experiment Environments and Settings}
We conducted our experiments on a Windows-based machine equipped with an Intel Xeon KVM CPU@2.9GHz and 16G of memory. Our model was a two-layer neural network consisting of 300 hidden ReLU neurons, which we trained on the MNIST dataset. The MNIST dataset comprises 10 labels, and contains 60,000 images for training and 10,000 images for testing. Each image is composed of 28 × 28 features (pixels) and represents a single digit ranging from 0 to 9. We applied L2 regularization with a rate of 0.001 and utilized a decaying learning rate strategy. Specifically, we set the learning rate to 0.2 for the first half of the training iterations and reduced it to 0.1 for the remaining iterations. We also utilized deterministic gradient descent as our optimization algorithm.

Regarding our hyperparameter configuration, 
we set $T_0=5$ as the period of explicit gradient updates, 
and $j_0=10$ as the length of the initial ``burn-in''. 
Specifically, 
for the two-layer DNN, 
we set $T_0=2$, which is even smaller, 
and we use the first quarter of the iterations as the ``burn-in''. 
The history size m is set to 2 for all of our experiments.

In the following experiments, 
we will assess the model's performance by varying the deletion ratios, 
which are set to 0.001, 0.005, 0.01, 0.05, 0.1, 0.2, and 0.3, respectively.
Furthermore, 
we study the effect of machine unlearning on data samples deleted in a non-IID manner. 

\subsection{Experimental Results}

This subsection presents our primary evaluation results, 
including accuracy evaluation in both IID and non-IID settings, 
as well as the inversion of time consumption. 
The models examined in this evaluation are as follows:

\begin{itemize}
    \item Baseline Training: training a model using the whole dataset (in Red color).
    \item Retraining from scratch: retraining a model using the updated dataset after deletion (in green color).
    \item Deltagrad: retraining a model using the deleted samples (in blue).
\end{itemize}

\subsubsection{Evaluation of accuracy in an IID setting}

In order to evaluate the accuracy of the model under an IID setting, we randomly removed data samples using a uniform distribution. Fig.~\ref{fig_iid} illustrates that the testing accuracy of all three models remains consistent throughout this process. As a result, we can conclude that the baseline model's accuracy is not affected by this form of data deletion.
However, 
for the other two models, 
the degradation in testing accuracy is limited, 
decreasing from approximately 0.8735 to 0.8725 (less than 0.1\%).

\subsubsection{Evaluation of accuracy in a non-IID setting}
\label{subsubsec:non-IID}
To evaluate the model accuracy in a non-IID setting, 
we deleted more samples with the label ``4'' than the other labels. 
In more detail, 
30\% of the deleted samples are with the label ``4'', 
while the remaining 70\% of the deleted samples have a uniform distribution over the 9 labels without ``4''. 
In this non-IID setting, 
the performance degradation of two ``unlearn'' models is quite significant. 
As shown in Fig.~\ref{fig_noniid}, 
the accuracy degradation of Deltagrad is from around 0.8735 to 0.7921 (over 8\%) when the deletion ratio increases to 30\%. 
This is mainly caused by the significant absence of label ``4''. 
Noticeably, 
the classification accuracy of Label “9” increases from 82.9\% (deletion ratio: 0.001) to 88.3\% (deletion ratio: 0.3).

\begin{figure}
\centering
\includegraphics[width=3.2in]{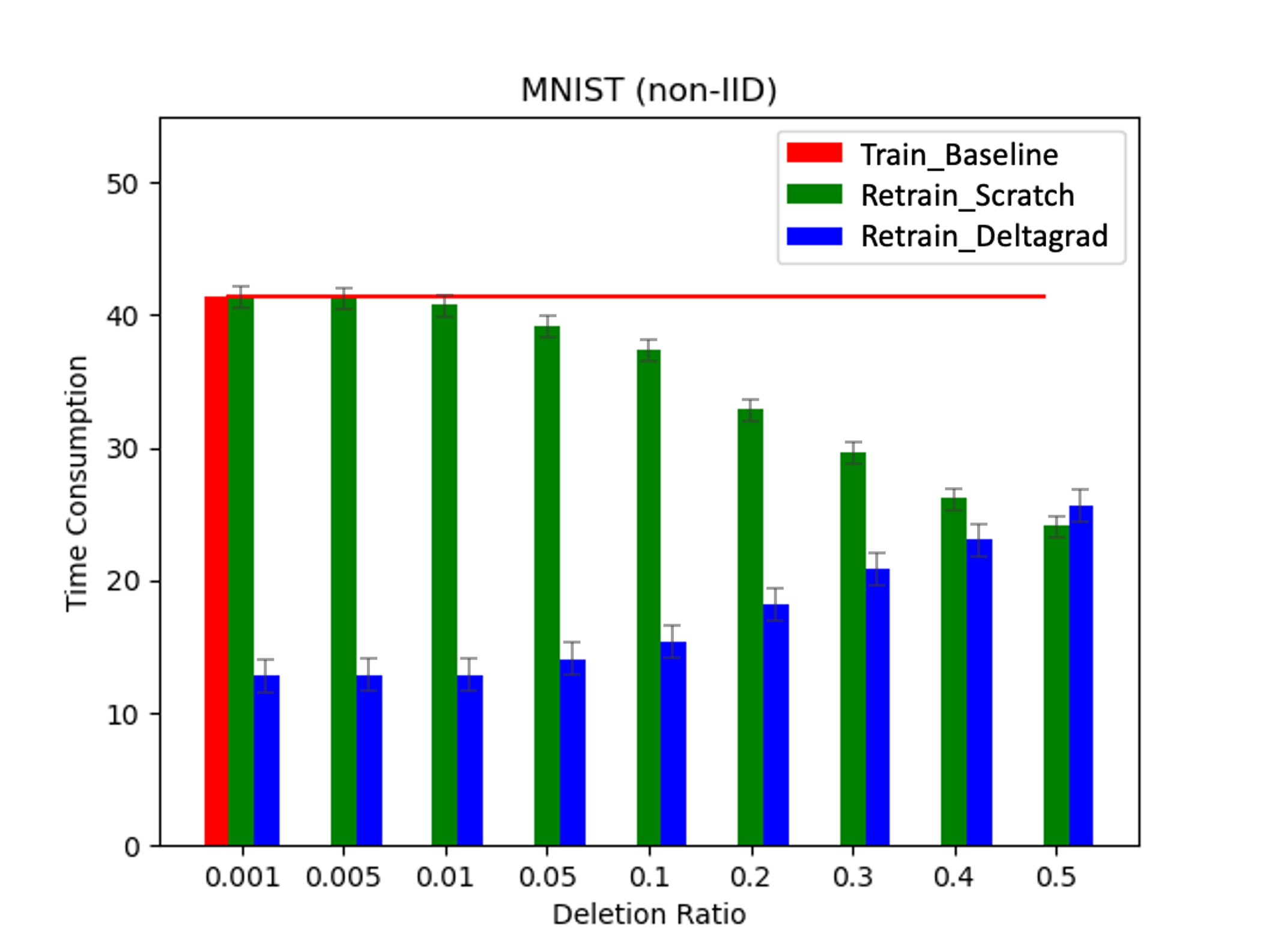}
\caption{Evaluation of time consumption}
\label{fig_time}
\end{figure}

\subsubsection{Evaluation of time consumption}

In Fig.~\ref{fig_time}, 
the time consumption v.s. deletion ratio in an IID setting is displayed. 
In particular, 
to show the crossover point, 
we evaluate two more deletion ratio values, 
which are 0.4 and 0.5. 
With the increase in the deletion ratio from 0.001 to 0.3, 
the time consumption of ``Re-training from scratch'' decreases from 42.19s to 29.15s. 
Meanwhile, 
the time consumption of ``Deltagrad'' increases from 12.53s to 21.22s. 
At last, 
there will exist a cross-over point. 
It is worth mentioning that the time consumption trend in a non-IID setting is almost the same.

\section{Insights and Challenges}
\label{sec::future}

This section shares some insights on the state of machine unlearning based on our analysis
and evaluation, and discusses further challenges and potential approaches.

\subsection{Insights}

\textbf{Training from scratch or not?} 
Given the negligible degradation of accuracy comparing training-from-scratch with Deltagrad in the IID setting, 
we can conclude that the machine unlearning methods should be selected according to other performance metrics such as computational complexity.

\textbf{Non-IID machine unlearning is challenging.} 
It has been widely acknowledged that non-IID data distribution in distributed learning paradigms,
such as federated learning, 
leads to suboptimal model performance. 
In the context of unlearning, 
the removal of non-IID samples also results in a tremendous performance loss, 
which must be taken into consideration in practice.

\textbf{Unlearning may lead to Unfairness.} 
When data samples are erased, 
particularly in a non-IID manner, 
the classification accuracy of certain labels can significantly degrade while the accuracy of others may improve. 
This phenomenon would cause unfairness across different classes, 
which cannot be revealed by traditional ML performance metrics such as average accuracy.

\textbf{Unlearning calls for novel and more generalized loss function formulation methods.} 
The majority of the present machine unlearning methods rely on maximizing a quadratic loss function using Newton's method. 
Nonetheless, 
in practical scenarios, 
various forms of loss functions exist. 
Exploring the uses of these alternatives may lead to the design of  
more adaptable machine unlearning mechanisms in the future.

\subsection{Challenges}

\textbf{Conceptual reasons why non-IID data can complicate unlearning:}

- Data Heterogeneity: In a non-IID setting, certain data points or classes may be overrepresented in the data that are to be unlearned. This can skew the model's understanding, particularly if the unlearning process inadvertently strips away vital information about underrepresented classes.

- Model Stability: When training on IID data, models often converge to a stable set of weights. With non-IID data, especially during unlearning, the model's weights can oscillate or diverge due to the uneven distribution of the deleted data. This can make the unlearning process unpredictable.

- Loss Landscape: Non-IID data can introduce sharp, non-convex regions in the loss landscape. When unlearning, navigating this complex terrain can lead to getting stuck in sub-optimal local minima or experiencing larger jumps in loss.

\textbf{Potential mitigations and areas of future research to address this issue:}

- Regularization Techniques: Implementing regularization methods, like dropout or L1/L2 regularization, may help in ensuring that the model does not overfit to the non-IID nature of the data being removed.

- Stratified Unlearning: Similar to stratified sampling, one could look into stratified unlearning, where data is removed in a way that maintains the original distribution of the dataset as much as possible.

- Meta-Learning: Future research could explore meta-learning, where models are trained not just on the task at hand but also on how to forget or unlearn effectively.

- Dynamic Learning Rates: Adjusting learning rates dynamically during the unlearning process based on the nature of the data being removed could be beneficial. This might help in handling the complexities arising from non-IID data removal.

\textbf{Broader implications for models that frequently undergo updates and deletions, especially in evolving and dynamic environments:}

- Continuous Re-evaluation: Models in dynamic environments should be re-evaluated regularly. This is crucial because, with every unlearning and retraining cycle, the model's performance may change unpredictably due to the non-IID nature of updates.

- Scalability Concerns: As models grow in size and complexity, the overhead of frequently updating and deleting becomes a challenge. Ensuring that the unlearning process is efficient and scalable is crucial.

- Privacy Concerns: In some cases, unlearning is driven by the need to remove sensitive or personal information. With non-IID data, ensuring complete removal becomes complex, leading to potential privacy breaches.

- Model Robustness: If a model is trained on a diverse set of data and then frequently updated with non-IID data or has non-IID deletions, its robustness can be compromised. This can make the model more susceptible to adversarial attacks or produce unexpected outputs in unfamiliar scenarios.

\section{Summary}
\label{sec::summary}

In summary, 
machine unlearning holds great promise for addressing concerns of privacy and equity in machine learning. 
In this short survey article, 
we have presented a comprehensive examination of current machine unlearning techniques, 
including both exact and approximate unlearning methods, 
and compared various techniques. 
The merits and shortcomings of each method have been discussed, 
and our experimental evaluations have demonstrated that machine unlearning can remove sensitive data or biases from a trained model without significantly impairing its performance, 
particularly in IID settings. 
However, 
there exist significant hurdles to overcome, 
such as developing robust methods for non-IID sample deletion and evaluating the effectiveness of machine unlearning techniques against potential attacks. 
We believe that further research in this field will continue to advance the state-of-the-art in machine unlearning, 
allowing for the development of more robust and trustworthy machine-learning systems that meet the needs of diverse stakeholders.

\ifCLASSOPTIONcaptionsoff
  \newpage
\fi

\bibliographystyle{IEEEtran}
\bibliography{nm}

\begin{thebibliography}{10}
\providecommand{\url}[1]{#1}
\csname url@samestyle\endcsname
\providecommand{\newblock}{\relax}
\providecommand{\bibinfo}[2]{#2}
\providecommand{\BIBentrySTDinterwordspacing}{\spaceskip=0pt\relax}
\providecommand{\BIBentryALTinterwordstretchfactor}{4}
\providecommand{\BIBentryALTinterwordspacing}{\spaceskip=\fontdimen2\font plus
\BIBentryALTinterwordstretchfactor\fontdimen3\font minus
  \fontdimen4\font\relax}
\providecommand{\BIBforeignlanguage}[2]{{%
\expandafter\ifx\csname l@#1\endcsname\relax
\typeout{** WARNING: IEEEtran.bst: No hyphenation pattern has been}%
\typeout{** loaded for the language `#1'. Using the pattern for}%
\typeout{** the default language instead.}%
\else
\language=\csname l@#1\endcsname
\fi
#2}}
\providecommand{\BIBdecl}{\relax}
\BIBdecl

\bibitem{chien2023efficient}
E.~Chien, C.~Pan, and O.~Milenkovic, ``Efficient model updates for approximate
  unlearning of graph-structured data,'' in \emph{International Conference on
  Learning Representations}, 2023.

\bibitem{wu2020deltagrad}
Y.~Wu, E.~Dobriban, and S.~Davidson, ``Deltagrad: Rapid retraining of machine
  learning models,'' in \emph{International Conference on Machine
  Learning}.\hskip 1em plus 0.5em minus 0.4em\relax PMLR, 2020, pp.
  10\,355--10\,366.

\bibitem{voigt2017eu}
P.~Voigt and A.~Von~dem Bussche, ``The {EU} general data protection regulation
  ({GDPR}),'' \emph{A Practical Guide, 1st Ed., Cham: Springer International
  Publishing}, vol.~10, no. 3152676, pp. 10--5555, 2017.

\bibitem{cao2015towards}
Y.~Cao and J.~Yang, ``Towards making systems forget with machine unlearning,''
  in \emph{2015 IEEE Symposium on Security and Privacy}.\hskip 1em plus 0.5em
  minus 0.4em\relax IEEE, 2015, pp. 463--480.

\bibitem{guo2020certified}
C.~Guo, T.~Goldstein, A.~Hannun, and L.~Van Der~Maaten, ``Certified data
  removal from machine learning models,'' in \emph{Proceedings of the 37th
  International Conference on Machine Learning}, 2020, pp. 3832--3842.

\bibitem{nguyen2020variational}
Q.~P. Nguyen, B.~K.~H. Low, and P.~Jaillet, ``Variational bayesian
  unlearning,'' \emph{Advances in Neural Information Processing Systems},
  vol.~33, pp. 16\,025--16\,036, 2020.

\bibitem{bourtoule2021machine}
L.~Bourtoule, V.~Chandrasekaran, C.~A. Choquette-Choo, H.~Jia, A.~Travers,
  B.~Zhang, D.~Lie, and N.~Papernot, ``Machine unlearning,'' in \emph{2021 IEEE
  Symposium on Security and Privacy (SP)}.\hskip 1em plus 0.5em minus
  0.4em\relax IEEE, 2021, pp. 141--159.

\bibitem{brophy2021machine}
J.~Brophy and D.~Lowd, ``Machine unlearning for random forests,'' in
  \emph{International Conference on Machine Learning}.\hskip 1em plus 0.5em
  minus 0.4em\relax PMLR, 2021, pp. 1092--1104.

\bibitem{golatkar2020eternal}
A.~Golatkar, A.~Achille, and S.~Soatto, ``Eternal sunshine of the spotless net:
  Selective forgetting in deep networks,'' in \emph{Proceedings of the IEEE/CVF
  Conference on Computer Vision and Pattern Recognition}, 2020, pp. 9304--9312.

\bibitem{sekhari2021remember}
A.~Sekhari, J.~Acharya, G.~Kamath, and A.~T. Suresh, ``Remember what you want
  to forget: Algorithms for machine unlearning,'' \emph{Advances in Neural
  Information Processing Systems}, vol.~34, pp. 18\,075--18\,086, 2021.

\bibitem{ginart2019making}
A.~Ginart, M.~Guan, G.~Valiant, and J.~Y. Zou, ``Making ai forget you: Data
  deletion in machine learning,'' \emph{Advances in neural information
  processing systems}, vol.~32, 2019.

\bibitem{sommer2022athena}
D.~M. Sommer, L.~Song, S.~Wagh, and P.~Mittal, ``Athena: Probabilistic
  verification of machine unlearning,'' \emph{Proceedings on Privacy Enhancing
  Technologies}, vol.~3, pp. 268--290, 2022.

\bibitem{chen2021machine}
M.~Chen, Z.~Zhang, T.~Wang, M.~Backes, M.~Humbert, and Y.~Zhang, ``When machine
  unlearning jeopardizes privacy,'' in \emph{Proceedings of the 2021 ACM SIGSAC
  Conference on Computer and Communications Security}, 2021, pp. 896--911.

\bibitem{marchant2022hard}
N.~G. Marchant, B.~I. Rubinstein, and S.~Alfeld, ``Hard to forget: Poisoning
  attacks on certified machine unlearning,'' in \emph{Proceedings of the AAAI
  Conference on Artificial Intelligence}, vol.~36, no.~7, 2022, pp. 7691--7700.

\bibitem{shokri2017}
R.~Shokri, M.~Stronati, C.~Song, and V.~Shmatikov, ``Membership inference
  attacks against machine learning models,'' in \emph{2017 IEEE Symposium on
  Security and Privacy (SP)}, 2017, pp. 3--18.

\end{thebibliography}

\end{document}